# Soft Sensors and Process Control using AI and Dynamic Simulation[†]

Shumpei Kubosawa[1,2,‡], Takashi Onishi[1,2] and Yoshimasa Tsuruoka[1,3]

[1]*NEC-AIST AI Cooperative Research Laboratory, National Institute of Advanced Industrial Science and Technology, 2-4-7 Aomi, Koto-ku, Tokyo 135-0064, Japan*

[2]*Data Science Research Laboratories, NEC Corporation, 1753 Shimonumabe, Nakahara-ku, Kawasaki-shi, Kanagawa 211-8666, Japan*

[3]*Department of Information and Communication Engineering, Graduate School of Information Science and Technology, The University of Tokyo, 7-3-1 Hongo, Bunkyo-ku, Tokyo, 113-8656 Japan*

***Keywords:*** Soft Sensors, Reinforcement Learning, Dynamic Simulation, Systems Identification

During the operation of a chemical plant, product quality must be consistently maintained, and the production of off-specification products should be minimized. Accordingly, process variables related to the product quality, such as the temperature and composition of materials at various parts of the plant must be measured, and appropriate operations (that is, control) must be performed based on the measurements. Some process variables, such as temperature and flow rate, can be measured continuously and instantaneously. However, other variables, such as composition and viscosity, can only be obtained through time-consuming analysis after sampling substances from the plant. Soft sensors have been proposed for estimating process variables that cannot be obtained in real time from easily measurable variables. However, the estimation accuracy of conventional statistical soft sensors, which are constructed from recorded measurements, can be very poor in unrecorded situations (extrapolation). In this study, we estimate the internal state variables of a plant by using a dynamic simulator that can estimate and predict even unrecorded situations on the basis of chemical engineering knowledge and an artificial intelligence (AI) technology called reinforcement learning, and propose to use the estimated internal state variables of a plant as soft sensors. In addition, we describe the prospects for plant operation and control using such soft sensors and the methodology to obtain the necessary prediction models (i.e., simulators) for the proposed system.

## Preface

Artificial intelligence (AI) technologies, including machine learning, have been widely applied in practical applications, such as supporting the operation of chemical plants. Machine learning is a technology that leverages a substantial amount of data to extract and utilize the requisite knowledge for automating and optimizing processes such as classification, identification, prediction, and control. In modern chemical plants, a database called "historian" is connected to the distributed control systems of the plants, and the operating data are automatically recorded. Using recorded operating data is becoming common for machine learning applications.

In chemical plants, to stably and continuously produce products of a given quality, optimal control is required to address each situation and operational objective. Optimal control involves finding the optimal amount of manipulations resulting in the highest consistency to a given target state, thereby leveraging the prediction of future plant states. Predictions are necessary for optimal control and plant operations in general. For example, in proportional-integral-derivative control (PID control), a control law is constructed to decrease the residual between the set-point variable (SV) and process variable (PV) values using a simple prediction model wherein "decreasing the manipulated variable (MV) value consequently decreases the PV value". In model predictive control (MPC), also known as "advanced control," a prediction model created from data recorded on the historian is used to calculate the amount of manipulation required to achieve the target state.

Thus, control requires prediction. Precise prediction of future states requires the current actual state, specifically, the current PV values, because they are the starting point of the prediction. Some variables, such as temperature and flow rate, can be obtained immediately, whereas measuring other variables, such as purity and viscosity, is time-consuming and intermittent. Among the variables that are

---

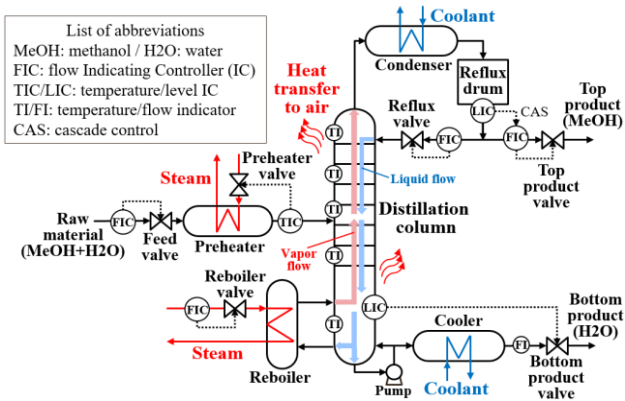

**Fig. 1** Overview (piping and instrumentation diagram) of a binary distillation plant (e.g., separating methanol and water from a mixture).

difficult to measure instantly, they may indicate product quality or strongly influence the maintenance of equipment, such as clogging. To maintain stable production by immediately responding to changing conditions, instantly acquiring these variables is desirable. Therefore, a "soft sensor" has been proposed to estimate variables that are difficult to measure immediately from variables that can be measured immediately. However, soft sensors constructed solely from historical data are limited by reduced estimation accuracy or extrapolation performance in situations in which no data are available.

In this study, we propose a method to construct soft sensors by combining reinforcement learning (RL), which is a branch of machine learning, and dynamic simulators that can estimate and predict states using chemical engineering principles such as physical laws and transport phenomena. The performance of the proposed soft sensor was evaluated in an actual plant, and the results indicated that the estimated and measured values were consistent. Additionally, we discuss the prospects for control guidance using the proposed soft sensors and those for obtaining the prediction models required for the proposed method.

## I. Background

Chemical plants utilize various physical and chemical phenomena, such as vapor–liquid equilibrium (evaporation) and reactions, to produce products. To precisely control the evolution of these phenomena and maintain stable production, chemical plants are equipped with numerous sensors and valves that extensively monitor and precisely operate the conditions. Although most of the installed sensors, such as thermometers, flowmeters, and pressure gauges, instantly indicate PV values, other variables, such as purity and viscosity, are difficult to measure instantly.

Plant operators estimate and predict internal states of the plant that are difficult to measure from instantly displayed sensor values. For distillation, which is a typical separation process, a relationship exists between the pressure, temperature, and composition (purity) of each stage of the distillation column. In a two-component distillation system (e.g., water and methanol (MeOH)) (**Figure 1**), at a constant column pressure, the purity of the low-boiling-point substance (MeOH) in the distillate (top product) increases with decreasing column top temperature, and that of the high-boiling-point substance (water) in the bottoms (bottom product) increases with increasing bottom temperature of the column. The operator estimates the composition of each stage based on these relationships and the temperature of each stage. If a change in product composition is predicted from the temperature changes in each stage, the operator adjusts the reboiler (heating rate) and reflux (cooling rate) to regulate the column conditions and maintain stable production by minimizing the impact on the product quality.

The internal states of a distillation column depend on various factors such as the flow rate and temperature of the raw material feed, reboiler, and reflux, and the external temperature, in addition to their temporal changes until the present time. The optimal operation also depends on the internal states. These properties are common to the operation of chemical plants besides distillation columns. Therefore, accurate estimations of the internal plant states and the calculation of appropriate optimal operations (control) based on future predictions are necessary; however, both require extensive knowledge of chemical engineering and experience or expertise in operation. By contrast, the number of skilled operators is limited, and training them takes time. To maintain stable production in such situations, most operators, regardless of their degree of expertise, are required to respond adequately. Therefore, automatic assistance in estimating the internal state, future prediction, and control using prediction is desirable.

## II. Related Work

### A. Soft sensor

Although some process variables can be measured immediately and continuously, others require offline analysis to obtain variable values and cannot be frequently measured. Although the variables are difficult to measure, they typically impact product quality, process control, and maintenance. Therefore, a soft sensor that estimates measurements by calculation has been proposed as a more frequent and economical method for measuring the variables. There are two types of soft sensors: model- and data-driven (Kadlec *et al.*, 2009). Model-driven methods generally leverage first-principles models, that is, mathematical models of physical phenomena. However, in conventional model-driven methods, models that focus on reproducing ideal steady-state conditions are commonly used. Thus, the model-driven method has limited use as a soft sensor that requires estimation of time-varying states. In contrast, for practical situations, data-driven methods for constructing soft sensors are widely employed, wherein soft sensors are constructed statistically from process data recorded by historians. In the data-driven methods, linear models, such as principal component analysis and partial least squares regression, or nonlinear models, such as neural networks and support vector machines, are used as estimation models.

*B. Dynamic Simulation*

Systems such as chemical plants, engines, and motors, whose output is determined by the immediate input and past input history, are called dynamical systems. In dynamical systems that utilize physical laws, the laws (physical models) of time-varying internal states are often described by differential equations. Dynamic simulation is used to reproduce and predict the time evolution (change) of the internal state and output of a dynamical system based on mathematical models of physical laws. The internal states and outputs are computed by integrating and solving the differential equations of state transition rules over time using the initial state and a series of system inputs. In chemical plants, dynamic simulators are widely employed to train operators to regulate non-steady-state conditions, such as startup, shutdown, or abnormal conditions (Kano and Ogawa, 2010).

*C. State Observer*

The internal states of a dynamical system may not be observable externally. For instance, the gas-phase composition of each stage of a distillation column cannot be directly observed, because a sample cannot be conveniently obtained. However, in some cases, using these variables as controls is desirable. State observers (**Figure 2**) are used in these cases. State observers use prediction models to reproduce the behavior of plants. Because the prediction model calculates the internal state, the state variables can be consequently obtained. The same manipulation values are input into this prediction model simultaneously with the plant. The corresponding outputs are then obtained, and the inputs and parameters of the prediction model are adjusted to ensure that the output of the prediction model is consistent with that of the plant. If the outputs of the prediction model are consistent with those of the plant after the adjustment, the internal states on which the outputs depend are also considered consistent. These state variables can then be obtained using the prediction model. State observers are a common technique used in control engineering and dynamic simulations.

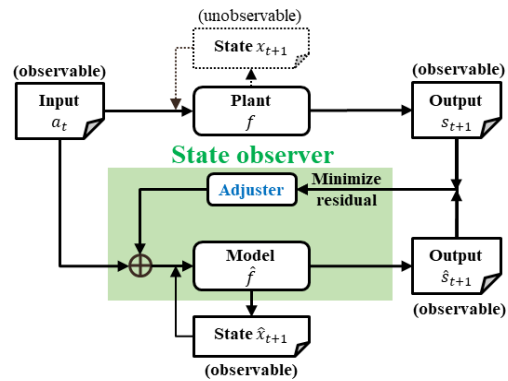

**Fig. 2** Schematic of a state observer that estimates unobservable inner states of the plant.

The method of using dynamic simulation for the prediction model of the state observer is called tracking simulation. To ensure that the simulation outputs correspond with those of the plant, the simulation parameters are automatically adjusted by PID control based on the residuals of each output (Nakaya and Li, 2013). The internal states estimated using this method can be used as the initial state of the simulation for subsequent predictions.

*D. Reinforcement Learning*

Reinforcement learning (RL) is a method for automatically constructing the optimal manipulation of a dynamical system to achieve a specified goal, that is, the optimal control rules, using machine learning. Although extensively studied, AlphaGo (Silver *et al.*, 2016), a Go game AI that uses RL, attracted renewed attention after it defeated a professional human Go player in 2015. Moreover, Go is a dynamical system in which the board state (observed values) changes over time according to the rules of the game, as the player and opponent place a Go stone (action) on the board. A discrete-time dynamical system such as Go can generally be formulated as a state transition probability

$$\hat{f}: S \times A \times S \to [0,1] \quad (1)$$

where $\hat{s}_t \in S$ is the output of the system (observed values) at time t and its space; $a_t \in A$ is the input of the system (action values or operation values) and its space. The next state is drawn from the probability distribution $P_{\hat{f}}(\hat{s}_{t+1}|\hat{s}_t, a_t)$. The unobservable states are assumed to be included in the output. For this dynamical system $\hat{f}$ and initial state $\hat{s}_0$, RL is used to solve the following optimization problem (Levine, 2018)

$$\operatorname{argmax}_\pi \sum_{t=0}^{\infty} \mathbb{E}_{a_t \sim P_\pi(a_t|\hat{s}_t)}[\gamma^t r(\hat{s}_t, a_t)] \quad (2)$$
$$\text{subject to } \hat{s}_{t+1} \sim P_{\hat{f}}(\hat{s}_{t+1}|\hat{s}_t, a_t)$$

where
$$r: S \times A \to \mathbb{R} \quad (3)$$
$$\pi: S \times A \to [0,1]. \quad (4)$$

$\gamma \in [0,1)$ is the discount factor, indicating that future rewards are highly discounted. It is a coefficient that essentially prevents the sum from diverging to infinity because its terminal time is infinite. $\mathbb{R}$ is a set of real numbers; $r$ is a real-valued function representing the evaluated values of states and actions (large values are desirable), and is called the reward function. By defining the reward function, the user can define the optimization problem, that is, the task (control problem) to be solved and the desired state. $\pi$ is the policy function that represents the probability $P_\pi(a_t|s_t)$ of selecting each action (manipulation) for each state. RL involves determining the policy distribution in which the action yielding the largest total future reward has the largest probability of selection in each state. For training, a set of data

$$D = \{(\hat{s}_k, a_k, r(\hat{s}_k, a_k), \hat{s}_{k+1}, a_{k+1})\}_{k=0}^{K} \quad (5)$$

consisting of system input/output and reward function values is used. $K$ denotes the number of sample instances. In RL, the system is manipulated to collect training data $D$ based on the policy function during training; the parameters of the policy function are updated (improved) using the collected data, and these procedures are repeated. Software that implements RL can be regarded as an autonomous agent (**Figure 3**) that operates the system based on its own decisions (policy function) in response to the observed values of the system and independently improves the policy function. **Figure 4** illustrates the fundamental concept of RL. Each state and action are evaluated by tracking future rewards back to the past state, and the policy is improved to obtain higher rewards in the future. In the example shown in the figure, the highest reward is obtained by continually selecting action $a_C$. Therefore, the desired behavior can be achieved by adjusting the parameters of the policy function to increase the probability value $\pi(s_A, a_C) = P_\pi(a_C|s_A)$ of selecting action $a_C$ in state $s_A$ and decrease the probability value $\pi(s_A, a_A) = P_\pi(a_A|s_A)$ of selecting action $a_A$.

Simulators are often used for training agents instead of actual systems that need to be manipulated. This is because RL is a machine-learning method based on data, that is, experience; thus, the agent may output inappropriate manipulations in an inexperienced (untrained) state. In particular, during training, the agent frequently encounters such situations, which may lead the system to an unexpected state owing to inappropriate manipulations by the agent. Therefore, agents are often trained using simulators or other environments that do not affect reality, and they can experience diverse situations before controlling actual plants.

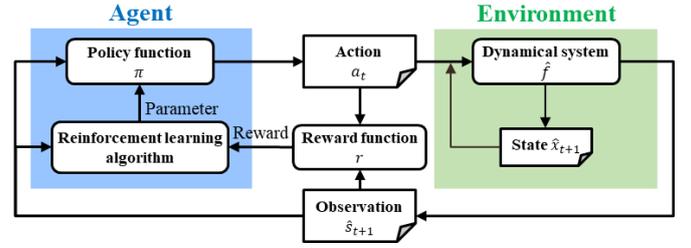

**Fig. 3** Schematic of reinforcement learning agent and target system.

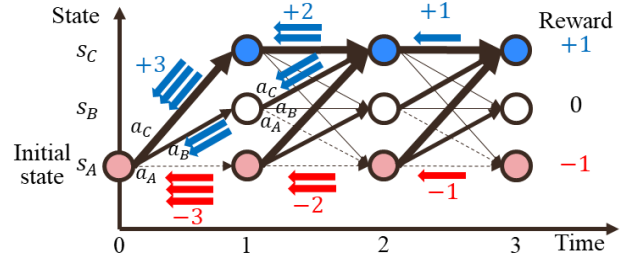

**Fig. 4** Basic concept of the reinforcement learning algorithm. Circles, black arrows, and colors depict states, actions, and rewards, respectively. The width of the action arrows indicates the selection probability of each action of the state. Future rewards are tracked back to past states (colored arrows) and used to evaluate states and actions.

To create various situations in the simulator, domain randomization is a proposed method, in which the simulation parameters (e.g., air temperature and feed composition) are randomly changed during training (Tobin *et al.*, 2017). In actual plants, reproducing a wide variety of situations is not practically feasible; thus, simulators are often used for training to avoid interrupting stable production. A well-trained policy function can provide optimal control in various situations.

*E. Model Predictive Control*

In addition to RL, optimal control methods include model predictive control (MPC), which is considered an advanced process control method. For a deterministic discrete-time dynamical system $\bar{f}$

$$\bar{f}: S \times A \to S \quad (6)$$

and an initial state $\hat{s}_0$, MPC is a method to solve an optimization problem

$$\operatorname{argmin}_{a_0} \sum_{t=0}^{N} \ell(\hat{s}_t, a_t) \text{ subject to } \hat{s}_{t+1} = \bar{f}(\hat{s}_t, a_t) \quad (7)$$

where
$$\ell: S \times A \to \mathbb{R} \quad (8)$$

to repeatedly obtain optimal manipulation at the next time

($a_0$) during control without prior training (Grüne and Pannek, 2017). $N$ is a natural number representing a finite number of periods, and this sum does not diverge from the definition; $\ell$ is a real-valued cost function that represents the evaluated value (small values are desirable) of the state and the manipulation. The cost function in MPC is often defined separately for stage and termination costs; however, if $N - t$ is included in the states, the termination cost can also be expressed by the above definition; thus, generality is not lost. Similar to the reward function, the user can define an optimization problem by defining a cost function.

The optimization problems of RL and MPC both involve maximizing or minimizing user-defined reward or cost, respectively. Therefore, the optimization problem of MPC can also be solved using RL. For example, in MPC, actions resulting a smaller value of cost function represents approaching the target state, and in RL, actions resulting a larger value of reward function represents approaching the target state. Both MPC and RL would output the same amount of manipulation for the subsequent time in the optimal solution (unless they are affected by the difference in the time period). Particularly, RL and MPC are two different approaches (pre-optimization and online optimization) to solve similar problems. However, RL handles the system stochastically and, as described below, does not set an upper bound on the optimization period; therefore, it can handle a wider range of problems than that handled by MPC.

In addition to the approach, other major differences are observed between RL and MPC. In MPC, the optimized period is $N$, and optimal manipulations are limited to the state up to $N$ times in the future. The restriction to $N$ can be considered a constraint to solve the optimization problem online. Specifically, the optimization problem is reduced in size by limiting it in terms of time, thereby reducing the amount of online computation at the expense of optimality in the future beyond $N$. In the analogy of game AI such as Go and Shogi, MPC optimizes the next move by "predicting $N$ moves ahead," and a larger $N$ would typically increase optimality; however, there is a trade-off in the waiting time, or computation time, which is triggered by the larger $N$. In contrast, for RL, there is no limit to the time period in which the algorithm can be optimized. In addition, the optimization computation is completed during prior training. Considering its application in Go, RL selects the optimal move (in a short time) based on intuition obtained from experience (training) with numerous long games. AlphaGo won against a professional Go player by combining the evaluation functions created by RL with a look-ahead search (Monte Carlo tree search).

RL is evidently capable of learning complex problems, such as Go, with sufficiently powerful computers in recent years. However, assessing the feasibility of learning individual tasks in a practical period remains challenging. Even in the same plant, when the reward or task settings are changed, the new setting is often impossible to train. From the perspective of a methodology for constructing an automatic controller, RL is limited by a large degree of uncertainty. In contrast, MPC is a realistic approach for constructing a controller that can reliably compute the solution and manipulation amount by reducing the size of the optimization problem, for example, by limiting the optimization period or restricting the prediction model to be linear. Owing to these differences in characteristics, MPC is used for problems that can be adequately represented by a linear model, whereas RL is used for those with strong nonlinearities. Regarding uncertainty of the construction in RL, it is possible to avoid "the curse of dimensionality," in which the number of candidate manipulations exponentially increases with the increasing number of manipulation points, making learning difficult by restricting the manipulation points of an agent, thereby reducing the uncertainty (Kubosawa *et al.*, 2021a).

*F. Plant Operation Guidance using RL and Simulator*

A method that uses RL and dynamic simulators to guide transition operations in a chemical plant to change the production rate or product composition has been proposed previously (Kubosawa *et al.*, 2021b, 2022). In large-scale continuous plants, MPC can be introduced to automatically respond to disturbances in the proximity of stable production at steady state. However, advances in automation and optimization are inadequate for transition operations, which require responses to non-steady states with various state changes and are thus difficult to predict and control. Therefore, RL has been adopted in previous studies for optimally controlling transition operations, which includes systems with complex state changes, that is, highly nonlinear systems. For example, in this method, when an operator specifies a target value for the purity of the top product in a distillation column, the guidance system provides an optimal operation plan based on the current plant state, continues to provide corrections for unexpected changes during the operation, and achieves the target product purity in the actual plant.

In previous studies, a method that combines three agents to achieve an optimal operation in such a transition operation has been considered (**Figure 5**). The first agent estimates the current internal states of the plant as a state observer and generates the initial states for the simulation.

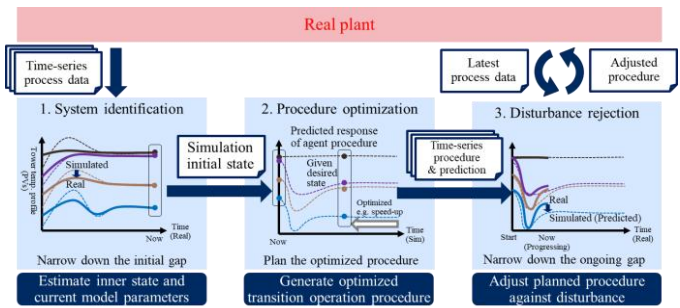

**Fig. 5** Process flow of transition operation guidance system for chemical plants (Kubosawa *et al.*, 2022). The control task is split into three subtasks; each subtask is addressed by the corresponding agent.

The second agent applies the estimated initial state as a starting point and creates an optimal operation plan to achieve the target state specified by the operator using the simulator to make predictions. The operation is then started according to the plan; however, the output of the actual plant may not appear as predicted owing to disturbances such as heavy rain, change in feed composition, other disturbances, or modeling errors. Therefore, the third agent corrects the initially planned manipulation values online based on the measured values of the actual plant to decrease the gap between the predictions and the actual plant. Combining these agents enables an actual plant to achieve its target state. This method is a framework that uses RL to manage real plants that are not always simulated.

In the previous study, experiments were conducted in an actual methanol (MeOH) distillation plant for training human operators (subsequently referred to as "M plant") operated by the Plant Operation Technology Training Center of Mitsui Chemicals, Inc. Figure 1 depicts the process flow of the M plant. An operator training simulator (a simulation model built on OmegaLand, which is an integrated simulation environment provided by Omega Simulation Corporation) installed at the plant was used in the experiments. The product purity was altered during the operation. The distillation column temperature was suddenly decreased by water spraying, which reproduced the disturbance caused by heavy rain. However, the operation could be adequately performed as planned with correction.

### III. PROPOSED METHOD

*Plant Operation Guidance using RL and Simulator*

The agent used in this study was the first agent used in a previous study on transition guidance. In particular, the method adjusts parameters using RL in a state observer that uses a dynamic simulator as a prediction model. We propose a method that uses this state observer as a model-driven soft sensor.

A previous study estimated the internal state of nonlinear dynamical systems using RL (Morimoto and Doya, 2007). As mentioned in the literature, RL can be suitably used for estimating the internal state from the observable output. There can be a time delay between a change in the internal state (parameters) of a dynamical system and that in its output. In this case, evaluating the parameter manipulation using the consistency of the outputs involves the values manipulated in the past time. Thus, parameter estimation is a delayed reward optimization problem wherein the current parameter values are reflected in future rewards. RL can address these problems, because it can maximize rewards in the future. Based on this previous study, the contribution of this study is threefold. First, by introducing automatic adjustment of rewards, we address possible anomalies in sensor indication values that may occur in real equipment and thus further improve the efficiency of training. Second, we demonstrate the effectiveness of using a dynamic simulator that comprehensively models a complex chemical plant, wherein discrete state changes (e.g., switching of cascade control) can occur. Third, the effectiveness of the method as a soft sensor in the real world is demonstrated through experiments using actual plant equipment.

In this method, a dynamic simulator is used for the model and an RL agent for the Adjuster in Figure 2 to adjust simulation parameter $a_t$. During training, control inputs $a_t$ and outputs $s_{t+1}$ recorded in the historian of the actual plant are used, and the inputs and outputs of the actual plant are transferred online for use.

The agent manipulates the parameter vector $\hat{a}_t$ of the simulator to ensure that the output of the simulator is consistent with that of the actual plant. Additionally, the same manipulation $a_t$ is simultaneously input to the simulator with the actual plant. If $\hat{s}_t$, $s_t$ are considered the output value vectors of the simulator and the actual plant at time t, and $\hat{x}_t$ is considered the state variable vector of the simulator, the observation values of the agent can then contain variables such as $\hat{a}_{t-1}$, $\hat{s}_t$, $a_t$, $s_t$, $\hat{x}_t$, $\hat{x}_{t-1}$, $\hat{x}_{t-2}$, and $\hat{x}_{t-3}$.

For example, in previous studies (Kubosawa *et al.*, 2021b, 2022), parameter $\hat{a}_t$ manipulated by the agent was the heat transfer coefficient to the air of the distillation column (two variables corresponding to the top and bottom sections); the heat balance was adjusted by manipulating these parameters, and the column temperature profile (temperature of each stage) of the simulator was made consistent with that of the actual plant.

Thus, the objective of the agent is to minimize the residual between the predicted and measured values. The reward function of the agent can be defined as

$$r(s_t, \hat{s}_t) := -\sum_i \alpha^{(i)} |\hat{s}_t^{(i)} - s_t^{(i)}| \qquad (9)$$

where the i-th element of the vector is denoted by the subscript '(i)'. $s_t$ is the vector of values that change as the agent manipulates the simulator, and $\hat{s}_t$ is the vector of observed values recorded in the historian, namely, the target data (constant) of the agent. $\alpha^{(i)}$ is a non-negative real number, which represents the weight of reducing the residual of the i-th sensor. For example, if the indicated value of a certain sensor on a real plant is typically erroneous, the efficiency of training can be increased by setting a small weight for that sensor and focusing on other sensors that can improve consistency.

The weights $\alpha^{(i)}$ can be set manually; for example, a value

$$\beta^{(i)} := 1 - \frac{1}{K}\sum_{t=0}^{K} \frac{(\hat{s}_t^{(i)} - s_t^{(i)})^2}{\sum_j (\hat{s}_t^{(j)} - s_t^{(j)})^2} \qquad (10)$$

$$\alpha^{(i)} := \beta^{(i)} / \sum_j \beta^{(j)} \qquad (11)$$

can be defined and updated periodically during training; $K$ is the length of the training data. This method automatically adjusts the weights of sensors such that it is easier to reproduce the actual values; sensors with small residual $(\hat{s}_t^{(i)} - s_t^{(i)})^2$ are adjusted to be larger, and those with large residuals to be smaller. Consequently, the training proceeds to initially reproduce the sensor values for which the residuals are easily minimized, and finally, the weights of the sensors that are difficult to reproduce remain small. Otherwise, as the residuals of all sensors approach zero, all weights are equalized. If the reliabilities of the actual sensors are constant for all target data, the weight vector $\alpha$ converges to a certain value if the training converges. This method is effective for sensors with similar types (units), such as a set of thermometers or flowmeters and for sensors with similar ranges (fluctuation ranges) of measured values, because the automatic adjustment is based on the comparison of residuals between sensors. Using this method, for example, the weights of the thermometers in each stage can be automatically adjusted. In a two-component distillation column, the temperatures at the bottom (reboiler) and top approach their respective boiling points and can be easily reproduced; however, the temperatures in the middle stages are difficult to reproduce, because they depend on the composition of each stage.

Nevertheless, training can progressively increase consistency from the bottom and top, where they are easier to reproduce, to the middle stages. With reference to RL, this weight adjustment method can be regarded as a type of reward shaping (Ng *et al.*, 1999), in which the difficulty of the training problem is reduced by modifying the reward function and accelerating the training.

Actual data recorded in the historian and the data newly generated on the simulator by artificially changing the simulation parameters can be used for target data $s_t^{(i)}$ during training. Training on domain-randomized data in this manner allows us to build agents that can respond to novel situations that have not yet been recorded in an actual plant.

The internal states of the simulator estimated by the agent trained in this manner can be used as a soft sensor. The gas-phase composition at each stage of the distillation column is estimated online. Additionally, the agent-estimated values of variables that cannot be measured, such as the heat transfer coefficient to the air of the distillation column, can be regarded as soft-sensor-indicated values. For example, if a sudden increase in the heat transfer coefficient to air is estimated, and a sudden drop in temperature owing to external factors such as heavy rain is identified, the corresponding operation can be performed; otherwise, an abnormality in the equipment is suspected. In addition, changes in the feed purity in the distillation column significantly affect the internal states of the column in addition to the flow rates of the top and bottom products. Purities, which are generally difficult to measure instantaneously, can be estimated by training an agent to manipulate them. Thus, the internal states from the simulation calculations, fluctuations in the simulation parameters, and battery limit conditions can be estimated using this method and subsequently used as soft sensors.

However, the parameters that can be estimated using this method are limited to a set of orthogonal parameters, namely, those of the regular model. For a distillation column, both the offset of the stream flow meter (the shift of the zero point) of the reboiler, which affects the amount of heat input to the column, and the heat transfer coefficient to air, which affects the amount of heat output from the column, can be adjusted simultaneously. However, if both are concurrently changed, the physical interpretation of these values is no longer feasible. This is because although the same situation (heat balance) can be reproduced by adjusting either parameter, no suitable criterion exists to reasonably select one out of the multiple parameter settings to reproduce that situation. By contrast, variables with small interactions, such as the heat and mass balances or variables of each highly independent process, can be expected to estimate reasonable values that can be interpreted physically. By training variables that should be estimated instantly and continuously as the manipulation value of the agent $\hat{a}_t$, the estimated value of $\hat{a}_t$, in addition to the internal state, can be used as the indicated value of the soft sensor.

## IV. EXPERIMENTAL RESULTS AND DISCUSSION

### A. Target Plant

The performance of the proposed soft sensor was experimentally evaluated using an actual M plant, similar to that in previous studies (Kubosawa *et al.*, 2021b, 2022), in addition to a dynamic simulator that reproduced the M plant.

### B. Experimental Settings

In this study, the agent estimated the feed purity (aqueous MeOH solution) and the two variables of the heat transfer coefficient to the air of the distillation columns, as in previous studies. Specifically, parameter $\hat{a}_t$, which is manipulated by the agent, consists of three variables: heat transfer coefficients to the air of the distillation column (two variables, i.e., top and bottom) and feed purity. These parameters are sequentially estimated online in real time at 5-min intervals and simultaneously fed into the simulator, which also estimates the internal states at 5-min intervals. Thus, the effectiveness of this method as a soft sensor for feed purity is evaluated. In the M plant, feed purity is not measured continuously but is obtained by offline analysis after a sample is withdrawn from the plant pipes. In addition to the parameter estimation for the heat balance in the previous study, the performance of estimating the feed purity, which is associated with the material balance, is evaluated in the present study to demonstrate the ability of this method to calculate both heat and material balances, which are fundamental concepts in chemical engineering. To address the material balance, the target sensors to be equalized are the top product flow rate and the temperature at each stage of the distillation column. Ideally, the sum of the top and bottom product flow rates equals the feed flow rate in the material balance.

Three experiments (R1, R2, and S) were conducted. Experiments R1 and R2 used the proposed method. Experiment R1 used simulator-generated data for the target data, whereas Experiment R2 used both the data recorded in the historian of the actual machine and the data generated by the simulator. For comparing the methods, in Experiment S, we evaluated the performance of an estimation model $m$ of $\hat{a}_t = m(s_t)$, which estimates the simulation parameters from the output values of the actual plant by supervised learning applying the same simulation data as in Experiment R1. Model $m$ is a data-driven soft sensor based on simulation data. In Experiment S, the simulator was not used during the training or after deployment. This has the advantage of reducing online computation during use.

The policy functions in Experiments R1 and R2 and the prediction model in Experiment S used neural networks. In Experiments R1 and R2, PPO (proximal policy optimization) (Schulman *et al.*, 2017) was used as the RL algorithm. In experiment S, the models were trained using the gradient method with the objective function of minimizing the squared error between the predicted values and training instances. The simulation data used for training in Experiments R1 and S were generated using domain randomization, which abruptly alters the heat transfer coefficients to air, thereby simulating heavy rainfall and alters feed purity in the simulator. For evaluating each experiment, the actual plant data from the historian and the feed and top product purity data sampled and measured during the experiment were considered. The number of target data episodes (in this case, the number of days of operation) used for training and that of actual plant data episodes used for the evaluation are listed in **Table 1**. Evaluation episodes are not included in the training data.

### C. Results

**Figure 6** presents the results of feed purity estimation based on actual operational data obtained in a single day for evaluation and the predicted top purity based on the estimated feed purity. The heat transfer coefficient to air is also estimated but not considered here. **Figure 7** displays the simulation results based on the estimated results and the sensor values from the actual plant. In experiments R1 and R2, the objective is to match the values from the simulator (thin line) to those from the actual plant (thick line) in the sensor values in Figure 7 by estimating the feed purity. Three indicators are used to evaluate performance. The first is the mean squared error (MSE, (A) in Table 1) between the feed purity in the simulator-generated data for evaluation and that estimated by each agent. The second is the mean squared error between the feed purity estimated by each agent (blue line in Figure 6) and the measured sample values (blue dots in Figure 6) in the actual operation data for evaluation ((B) in Table 1). Considering the same actual plant data for evaluation as that used for the second indicator, the third is the mean squared error ((C) in Table 1) between the estimated top product purity (blue line in Figure 6), which is predicted by simulation using the feed purity estimated by the agent, and the measured sample value (blue dots in Figure 6). In all cases, lower values indicate higher performance with fewer errors. The experimental results imply that experiment R2, in which both real and simulated data are used as target data for training, demonstrates the highest performance for all measures, and experiment R1, in which only simulated data are used, exhibits the second highest performance.

**Table 1** Experimental settings and results for comparing performance between reinforcement learning (RL) and supervised learning

| Experiment | R1 | R2 | S |
|---|---|---|---|
| Method | Proposed (RL) | | Supervised learning |
| # of training episodes (simulator) | 200 | 50 | 200 |
| # of training episodes (real plant) | 0 | 25 | n/a |
| (A) MSE of feed purity on 5 evaluation episodes (simulator) | 12.15 | 9.54 | 82.75 |
| (B) MSE of feed purity on 1 evaluation episode (real plant) | 3.98 | 2.28 | 299.11 |
| (C) MSE of top purity on 1 evaluation episode (real plant) | 0.02 | 0.02 | 25.58 |

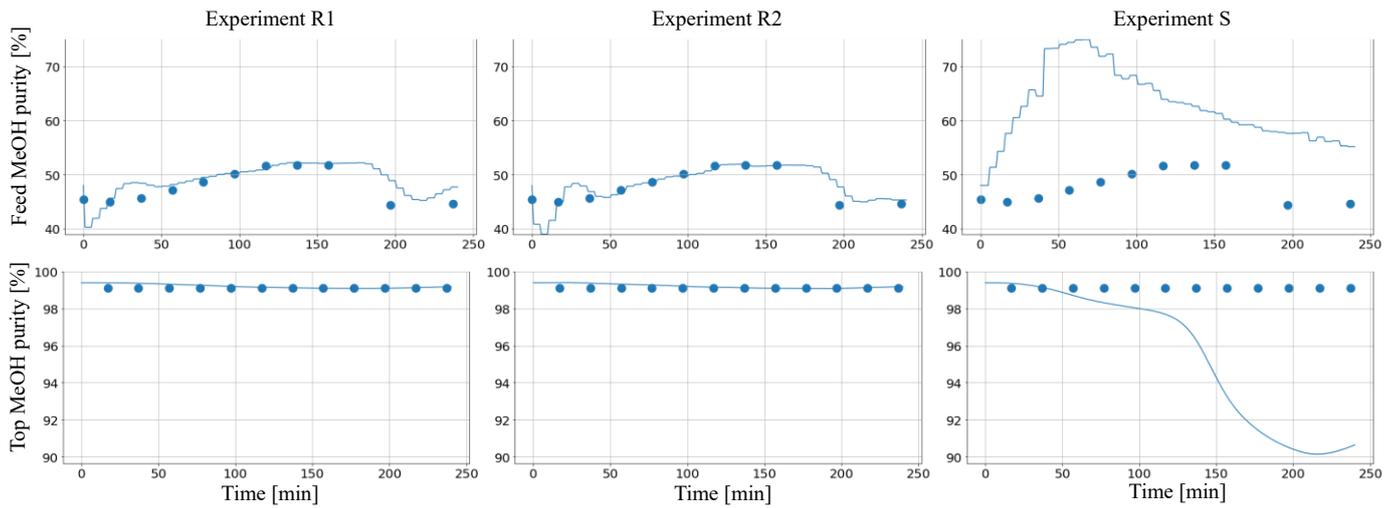

**Fig. 6** Estimated results of methanol purities in experiments R1 (left), R2 (center), S (right), and others. The top and bottom rows denote feed and top product purities in a day, respectively. Bottom product purities always approach zero and are therefore omitted. Solid lines and points indicate estimated and measured values, respectively.

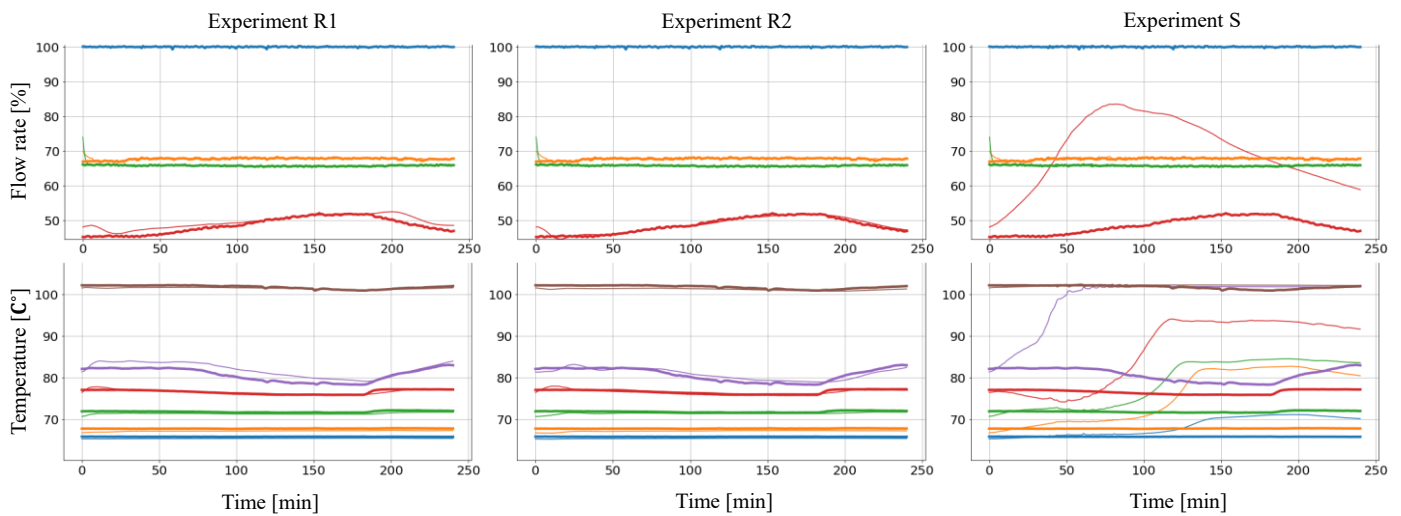

**Fig. 7** Simulated (thin) and actual (bold) flow rates (top row) and tower temperature profile (bottom row). Flow rates include feed (blue), reboiler steam (orange), reflux (green), and top product (red) flow in percent. The tower temperature profile consists of six thermometers: top (blue), bottom (reboiler) (brown), and middle stages (other colors).

In experiments R1 and R2, the estimated values are wavy at the beginning of the estimation (left side), implying a low estimation accuracy; however, after a certain duration, the estimation becomes highly accurate. This behavior is a common feature of state observers and is attributed to errors generated by starting the estimation from an unknown initial state. The M plant is not continuously operated for train-ing purposes, and the left end of the images in Figure 6 indicates the point in time immediately after starting up. The proposed method can continue to provide a highly accurate estimation during continuous operation; thus, it is suitable for practical use in continuous plants. In experiments R1 and R2, the error after the third measurement point is generally less than 2%. In both experiments R1 and R2, the larger value of the feed purity estimation error in the simulation data (A) than that in the actual feed purity estimation error (B) is also because of the estimation error at the beginning of the estimation period. At the beginning of the estimation, the estimated value demonstrates large fluctuations; however, in this experiment, the error is coincidentally small in the proximity of the actual sample measurements, and that between the actual sample measurements is large owing to the fluctuations in the estimated value.

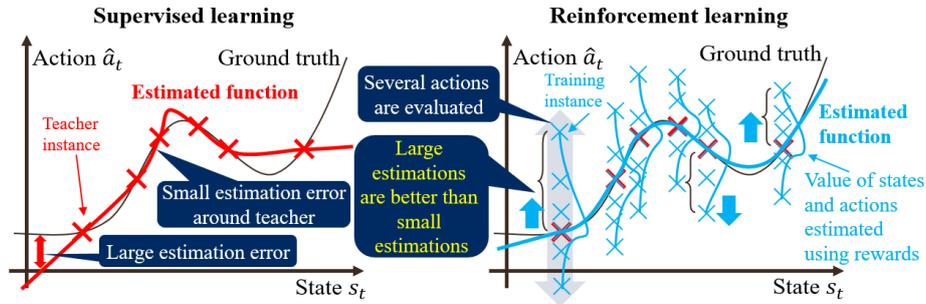

**Fig. 8** Schematic of supervised learning and reinforcement learning. Reinforcement learning indicates "the limitations of the estimation."

In addition to (B), the performance of R2 is better than that of R1 in (A). This is possibly because the inherent parameter variation trends in the actual plant data used in R2 better represent the characteristics of the process and disturbances than the trends in the domain randomization used in R1 while generating simulation data. Nevertheless, both R1 and R2 are comparable in performance, indicating that although simulation data can be independently used for training, using actual data improves the accuracy. Accordingly, additional training with new data accumulated over time can reflect recent trends and improve estimation performance. In this study, the training data were periodically added and re-trained.

In contrast to the results of experiments R1 and R2, the estimation error in experiment S is approximately 10% throughout the experiment. In Experiment S, the top product flow rate (thin red line in the top image of Figure 7) is higher than the measured value because the feed purity is estimated to be higher; moreover, the top product purity in Figure 6 is far from the measured value. This implies that the estimation model cannot adequately reproduce the material balance of an actual plant. Thus, the estimation model

(policy function) constructed using RL has a higher generalization (extrapolation) performance than that using supervised learning.

*D. Discussion: RL versus Supervised Learning*

The difference between RL and supervised learning in this research setting is that no error cases are observed in supervised learning. In supervised learning, "the distance between the output value of the estimated model and the target output value" is defined; however, "the validity of the output value" is not considered (**Figure 8**). The objective of supervised learning is to minimize the error between the output value and the target output only at points in the teacher data (marked with a red X in the figure), and nothing is considered for points away from the teacher data. However, in RL, various values are tested in the prediction model to obtain an evaluation value (reward), and future rewards are also reflected in the estimation. Thus, for example, insights such as "the apparent advantage of using a larger value instead of a smaller value in a certain situation" can be acquired. Particularly, the constraints and characteristics of the process (system), such as "the extent of deterioration of the evaluation value when the output value is slightly different from the supervised data," can be learned. Therefore, extrapolation ability of RL is considerably better than that of supervised learning. Although similar training is possible with supervised learning if a sufficiently large amount of training data is available, the advantage of RL is that the developer does not need to prepare "training data that appropriately represents process characteristics", because the agent can automatically and selectively collect training data by manipulating the environment.

V.  PROSPECTS OF AI AND SIMULATION

In this study, we propose a method that partially applies the method proposed in previous studies (Kubosawa *et al.*, 2021b, 2022) for system identification as a soft sensor, and its performance is evaluated. The applications of AI and simulators are not limited to soft sensors but also include control applications.

For example, the feed tank can be switched in response to changes in feed purity to minimize the impact on the product. Alternatively, as shown in previous studies, a variable estimated as a soft sensor, such as the top product purity of the distillation column, can be used as the target value for control. In the present operation, the target steady state at the target purity and SV values of each PID controller are first estimated by optimization on a static simulator, and the operator subsequently manipulates the SV values of the PID controllers to achieve the target steady state. Because considering complex dynamics is difficult when manipulating the SV value, monotonic manipulations, such as linearly changing the SV value from the present to the target final SV value, are often performed. In contrast, using AI and a simulator, optimal operation is possible by automatically finding the steady state of the target and by considering the dynamics during transition operations. A previous study (Kubosawa *et al.*, 2021a) revealed that when transition operations were optimized (trained) to minimize the time required for transition operations and the amount of steam and other utilities, both parameters in a production rate (load) change operation could be reduced by approximately half in contrast to a monotonic operation. The improved optimality assists operators and contributes to a sustainable society by reducing the environmental impact. As described previously, using AI and simulators enables operations involving complex dynamics and improves control optimality.

When using such a soft sensor for control, better control performance can be obtained if the sensor can rapidly detect state changes attributed to disturbances and other factors. In this study, in experiment R2, the estimated value requires approximately 5 min to track the measured value at a time point of approximately 200 min, when the measured sample value of the feed purity sharply declines. In a previous study (Kubosawa *et al.*, 2022, Figure 12), after a sudden drop in temperature owing to water spraying (disturbance) on the plant, approximately 5 min elapsed until the estimated heat transfer coefficient to air began to change significantly. In these experiments, the agent detected the change in approximately one step because the operation interval was 5 min. Therefore, the estimation was sufficiently prompt. Further reduction of the operation interval could potentially improve promptness. Additionally, the estimation is possible even with values that are not directly measured, such as the feed purity in this study, because the overall plant conditions, such as the temperature at each column stage, reflux flow rate, and product flow rates, are observed. Therefore, additional easily measurable sensors can ensure that the states of more locations are effectively obtained, thereby improving the promptness of the estimation.

VI. STRATEGY FOR OBTAINING PREDICTION MODEL

Any approach based on prediction and training, including this study, requires a simulator (prediction model). A dynamic simulator can be considered a "knowledge base for forecasting," which contains plant design information, physical laws, and chemical engineering principles; this simulator is desirable for forecasting non-steady-state situations. If an operator training simulator or another simulator is already available, it is relatively easy to obtain, otherwise a prediction model must be developed. Owing to the limited data available for non-steady states, creating a statistical model based entirely on data is impractical. Therefore, a strategy to obtain a prediction model is proposed in this section.

A. *Building Models while Designing the Plant*

The first approach is to embed the modeling of a dynamic simulator for a new plant in the design process of the plant. Thus, the optimal operation of the proposed plant design can be achieved through RL, and the performance of the plant, including its dynamics, can be quantitatively evaluated before construction. Furthermore, when establishing a standard operation procedure (SOP) after the design, operations can be quantitatively described, and an operation support AI can be constructed. This appears favorable; however, design engineers may specify from experience that simulators are not completely useful because they do not exactly reproduce reality and that they only add unnecessary work to the design process. However, as introduced in this study, RL methods have been proposed to address the gaps in practical application. Thus, these methods can open avenues wherein prediction and control can be simultaneously attained with the design.

B. *Symbolic Regression*

Symbolic regression is a type of analysis that has become popular since the 2010s. Symbolic regression is a method that improves prediction accuracy and ensures interpretability and extrapolation; both the prediction equation and the parameters of the equation are estimated from the data while preparing quadratic, exponential, trigonometric, and other functions as elements of the model (approximated function). Nevertheless, even if a model is constructed solely from data as a regression problem, accuracy in prediction or extrapolation of situations not recorded in the data cannot be expected. However, if prediction equations based on physical laws and chemical engineering principles are prepared as elements of the model, thereby the model candidates are limited in advance, the extrapolation ability of the data-based model can be improved. Examples of relevant applications in chemical process systems (Babu and Karthik, 2007) and materials science (Wang *et al.*, 2019) have been reported.

*C. Offline RL*

If a model is inevitably unavailable, using an actual plant as a model is the only remaining option. However, instead of connecting the agent to the plant during training, historical operation data can be used for RL. Generally, in RL, "training progresses as the agent manipulates the model"; however, the essence of RL is to "facilitate good operations that will yield more rewards in the future and inhibit bad operations that will not yield any rewards." Therefore, RL can also be conducted by evaluating the operating data recorded by a historian with an arbitrary reward function. This method is called offline RL (Levine *et al.*, 2020) because the agent and model are not connected online. Although learning operations that do not exist in the data is impossible, the best operation for the desired reward setting can be learned by reconnecting fragments of the recorded operations.

Offline RL can be achieved using online RL algorithms. For example, offline RL is possible with the PPO used in this study. For the PPO, the procedure in which the agent manipulates the simulator to obtain a response is simply switched to obtain the response from the historian. **Figure 9** depicts the results of the operations after online (top) and offline (bottom) RL. These are the results of the top product purity change runs, where the time evolution of the top product purity is evaluated in multiple episodes with different initial states and target purities, and the results are overlaid and plotted in blue lines. The target purity for each episode is plotted as an overlaid orange line. The faster the blue line overlaps with the orange line, the faster the agent achieves the target purity. The online algorithm can be used for training in an offline setting, although its performance is higher in the online setting.

However, for on-policy algorithms, such as PPO, the limitation is that the output value of the policy function and the operational value of the historian are significantly different in the early stages of training. If policy function

$$P_\pi(a_t|s_t) = \pi(s_t, a_t) \quad (12)$$

is represented by a Gaussian distribution $\mathcal{N}(\mu_{s_t}, \sigma^2_{s_t})$ with mean $\mu_{s_t} \in A$ and variance $\sigma^2_{s_t} \in [0, \infty)$, the policy function provides outputs $\mu_{s_t}$ and $\sigma^2_{s_t}$, depending on its input $s_t$. During training, the manipulation value is sampled from this policy distribution $a_t \sim \mathcal{N}(\mu_{s_t}, \sigma^2_{s_t})$ and tested on the simulator. By adding the noise defined by $\sigma^2_{s_t}$ through sampling, instead of the considered best manipulation $\mu_{s_t}$, the policy function can test "potential outcomes if other manipulation values are used," and search for a better option. After training, the optimal manipulation value of the training result can be obtained by selecting the manipulation value with the largest probability value, namely, $\mu_{s_t}$. Because the values of $\mu_{s_t}$ (and $\sigma^2_{s_t}$) are optimized by training, $\mu_{s_t}$ is often significantly different from that in the early and late stages of training. Similarly, the manipulation of historian $\tilde{a}_t$ (good manipulation) and that of the policy distribution at the beginning of training $\mu_{s_t}$ (untrained near-random values) can also be significantly distant. In offline RL, if the manipulation $\tilde{a}_t$ of the historian is regarded as "coincidentally sampled from the policy distribution" at the observed value $s_t$ of the historian, the historian data can be used for training. If $\tilde{a}_t$ and $\mu_{s_t}$ are distant from each other, the probability value $P(\tilde{a}_t|s_t)$ of $\tilde{a}_t$ in the policy distribution is extremely small (value at the edge of the Gaussian distribution; **Figure 10**). If this manipulation has a large value (expected future reward) and is updated to increase its probability value, then after a few times of updating policy parameters, the probability value of the manipulation is slightly increased (in this case, $\mu_{s_t}$ approaches $\tilde{a}_t$ and $\sigma_{s_t}^2$ would be widen. However, the increase is sharp when compared with the extremely small value $\log P_{\pi_{\text{old}}}(\tilde{a}_t|s_t) \ll 0$ of the policy distribution before the update (although it remains small). In this case, the probability ratio $P_\pi(a_t|s_t)/P_{\pi_{\text{old}}}(\tilde{a}_t|s_t) = \exp(\log P_\pi(\tilde{a}_t|s_t) - \log P_{\pi_{\text{old}}}(a_t|s_t))$ used in the PPO for updating parameters may overflow. Thus, if the probability ratio is extremely large, the update should be skipped. In the original on-policy case, updating the manipulation values sampled from the current policy distribution is unlikely to cause such large changes before and after the update. In addition to these adaptations, the hyperparameters and reward function settings must be revised to use the on-policy method offline. However, there is a significant engineering advantage to using a program intended for online training for offline training with almost no modification.

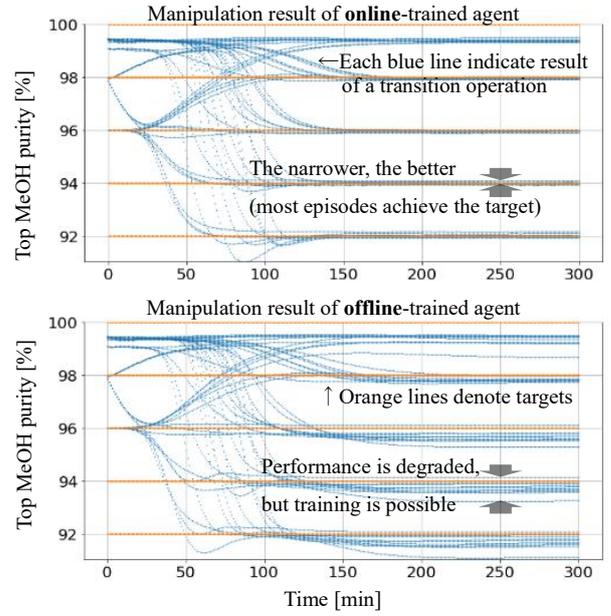

**Fig. 9** Time series of top product purity (blue) and target purity (orange) on the simulator, which is manipulated by online-trained (top) and offline-trained (bottom) agents. Multiple episodes are overlapped. Blue lines approach orange lines if each target is achieved.

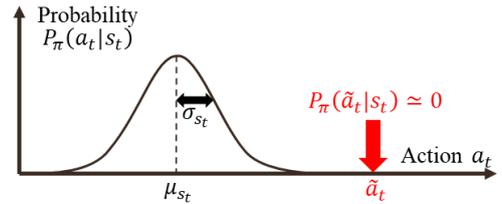

**Fig. 10** Schematic of policy distribution and historian action $\tilde{a}_t$.

SUMMARY

To construct a soft sensor, we proposed a method of combining RL and a dynamic simulator, which is a mathematical model of physical phenomena, and demonstrated its performance through experiments using both a simulator and an actual plant. Additionally, we proposed a method for operation control based on the values estimated by the soft sensor. In addition, we discussed the prospects of obtaining prediction models for constructing a soft sensor required for this method. In the future, we aim to further evaluate these methods in actual plants and improve their applicability.

ACKNOWLEDGEMENTS

Mitsui Chemicals, Inc. provided the authors with the opportunity to conduct experiments at the MeOH distillation plant for training at its Plant Operation Technology Training Center, and Omega Simulation, Inc. provided the authors with OmegaLand V3.2, an integrated dynamic simulation environment. Furthermore, both companies provided the authors with useful advice on the need for AI-based plant operation guidance and experimental setup. We would like to express our gratitude to them.

LITERATURE CITED

Babu, B. V. and S. Karthik; "Genetic Programming for Symbolic Regression of Chemical Process Systems," *Eng. Lett.*, **14**, 42–55 (2007).

Grüne, L. and J. Pannek; "Nonlinear Model Predictive Control." *Nonlinear model predictive control*, Springer, Cham, 45–69 (2017).

Kadlec, P., B. Gabrys, and S. Strandt; "Data-driven Soft Sensors in the Process Industry," *Comput. And Chem. Eng.*, **33**, 795–814 (2009).

Kano, M. and M. Ogawa; "The State of the Art in Chemical Process Control in Japan: Good Practice and Questionnaire Survey," *J. Process Control*, **20**, 969–982 (2010).

Kubosawa, S., T. Onishi, and Y. Tsuruoka; "Computing Operation Procedures for Chemical Plants using Whole-Plant Simulation Models," *Control Eng. Pract.*, **114**, 10487 (2021a).

Kubosawa, S., T. Onishi and Y. Tsuruoka; "Non-steady-state Control under Disturbances: Navigating Plant Operation via Simulation-Based Reinforcement Learning," 2021 60th Annual Conference of the Society of Instrument and Control Engineers of Japan (SICE), 799–806 (2021b).

Kubosawa, S., T. Onishi and Y. Tsuruoka; "Sim-to-Real Transfer in Reinforcement Learning-based Non-steady-state Control for Chemical Plants," *SICE J. Control Meas. Syst. Integr.*, **15**, 10–23 (2022).

Levine, S.; "Reinforcement Learning and Control as Probabilistic Inference: Tutorial and Review," *arXiv preprint*, arXiv:1805.00909 (2018).

Levine, S., A. Kumar, G. Tucker and J. Fu; "Offline Reinforcement Learning: Tutorial, Review, and Perspectives on Open Problems," *arXiv preprint*, arXiv:2005.01643 (2020).

Morimoto, J. and K. Doya; "Reinforcement Learning State Estimator," *Neural Comput.*, **19**, 730–756 (2007).

Nakaya, M. and X. Li; "On-line Tracking Simulator with a Hybrid of Physical and Just-In-Time Models," *J. Process Contr*, **23**, 171–178 (2013).

Ng, A. Y., D. Harada and S. Russell; "Policy Invariance Under Reward Transformations: Theory and Application to Reward Shaping," In Proceedings of the Sixteenth International Conference on Machine Learning, 278–287 (1999).

Schulman, J., F. Wolski, P. Dhariwal, A. Radford and O. Klimov; "Proximal Policy Optimization Algorithms," *arXiv preprint*, arXiv:1707.06347 (2017).

Silver, D., A. Huang, C. J. Maddison, A. Guez, L. Sifre, G. van den Driessche, J. Schrittwieser, I. Antonoglou, V. Panneershelvam, M. Lanctot, S. Dieleman, D. Grewe, J. Nham, N. Kalchbrenner, I. Sutskever, T. Lillicrap, M. Leach, K. Kavukcuoglu, T. Graepel and D. Hassabis; "Mastering the Game of Go with Deep Neural Networks and Tree search," *Nature*, **529**, 484–489 (2016).

Tobin, J., R. Fong, A. Ray, J. Schneider, W. Zaremba and P. Abbeel; "Domain Randomization for Transferring Deep Neural Networks from Simulation to the Real World," In the 2017 IEEE/RSJ International Conference on Intelligent Robots and Systems (IROS), IEEE, 23–30 (2017).

Wang, Y., N. Wagner and J. M. Rondinelli; "Symbolic Regression in Materials Science," *MRS Commun.* **9**, 793–805 (2019).